\documentclass[a4paper,fleqn]{cas-sc}

\usepackage[authoryear,longnamesfirst]{natbib}
\usepackage{lipsum}
\usepackage{soul}
\usepackage{subcaption}

\usepackage{}
\usepackage{placeins}
\usepackage{caption}
\usepackage{subcaption} 

\def\tsc#1{\csdef{#1}{\textsc{\lowercase{#1}}\xspace}}
\tsc{WGM}
\tsc{QE}

\begin{document}
\let\WriteBookmarks\relax
\def\floatpagepagefraction{1}
\def\textpagefraction{.001}

\shorttitle{}    

\shortauthors{}  

\title [mode = title]{Intelligent Spatial Estimation for Fire Hazards in Engineering Sites: An Enhanced YOLOv8-Powered Proximity Analysis Framework
}  

\tnotemark[] 

\tnotetext[1]{} 

%

\author[1]{Ammar K. AlMhdawi}

\cormark[1]

\fnmark[1]

\ead{}

\ead[url]{a.almhdawi@greatermanchester.ac.uk}

\credit{Conceptualization, Methodology, Software, Visualisation, Investigation, Formal analysis, Data curation, Writing -- original draft}

\affiliation[1]{organization={University of Greater Manchester},
            addressline={A676 Deane Rd}, 
            city={Bolton},
            postcode={BL3 5AB}, 
            state={},
            country={UK}}

\author[2]{Nonso Nnamoko}

\fnmark[2]

\ead{nnamokon@edgehill.ac.uk}

\ead[url]{}

\credit{Methodology,  Validation, Resources, Writing -- original draft}

\affiliation[2]{organization={Edge Hill University},
            addressline={St Helens Rd}, 
            city={Ormskirk},
            postcode={L39 4QP}, 
            state={},
            country={UK}}

\author[3]{Alaa Mashan Ubaid}

\cormark[1]

\fnmark[1]

\ead{alaa.ubaid@ukf.ac.ae}

\ead[url]{}

\credit{Formal analysis, Investigation, Writing -- review \& editing}

\affiliation[1]{organization={University of Khorfakkan},
            addressline={Al Jaradia }, 
            city={Sharjah},
            postcode={887W+2P2}, 
            state={},
            country={UAE}}


\cortext[1]{Corresponding author}

\fntext[1]{}


\begin{abstract}
\begin{abstract}
This study proposes an enhanced dual-based YOLOv8 framework for intelligent fire detection and proximity-aware risk assessment, extending conventional vision-based monitoring beyond detection to actionable hazard prioritization. The system is trained on a diverse dataset of 9{,}860 annotated images to accurately segment fire and smoke across complex visual environments. Central to the approach is a dual-model pipeline: a primary YOLOv8 instance segmentation model that detects and segments fire and smoke regions, and a secondary object detection model pretrained on the COCO dataset that identifies surrounding entities such as people, vehicles, and infrastructure components. By fusing the outputs of both models, the proposed method computes pairwise pixel-based distances between detected fire instances and nearby objects and converts these measurements into approximate metric units using a pixels-per-meter scaling procedure, enabling interpretable proximity estimation.

To operationalize spatial threat evaluation, the proximity information is integrated into a structured risk assessment model that combines fire evidence (segmentation confidence and spatial extent), object vulnerability (class-dependent consequence relevance), and distance-driven exposure decay to produce a quantitative risk score and corresponding alert tiers. The enhanced framework achieves strong performance, with precision, recall, and F1 scores exceeding 90\% and a mean average precision (mAP@0.5) above 91\%. The system outputs visually annotated frames and animated sequences showing fire locations, detected object categories, estimated distances, and risk-relevant context to support intuitive situational awareness and decision-making. Implemented entirely using open-source tools and executed within the Google Colab environment, the framework is lightweight, accessible, and deployable across diverse settings, including industrial sites and under-resourced environments where rapid response infrastructure may be limited. Overall, this work contributes a scalable and interpretable fire monitoring approach that links deep learning segmentation and object understanding with proximity-to-risk reasoning, providing a practical foundation for future integration with automated alerting, predictive analytics, and response systems in high-risk environments.
\end{abstract}




\maketitle

\section{Introduction}\label{sec:introduction}

Fire incidents remain one of the most destructive hazards in industrial, residential, and environmental contexts, causing significant loss of life, infrastructure damage, and economic disruption worldwide. Early and reliable fire detection is therefore critical for effective emergency response and risk mitigation. Traditional fire detection systems, including smoke sensors and thermal alarms, often suffer from delayed response times, limited spatial awareness, and high false alarm rates \citep{celik2007fire}. 

With the advancement of computer vision and deep learning, convolutional neural networks (CNNs) have demonstrated remarkable performance in object detection and segmentation tasks \citep{ren2015faster, redmon2016yolo}. In particular, the YOLO (You Only Look Once) family of detectors has enabled real-time object detection with high accuracy and computational efficiency \citep{redmon2016yolo, redmon2018yolov3}. More recently, YOLOv8 has introduced architectural improvements that enhance instance segmentation and detection robustness in complex environments \citep{jocher2023yolov8}. These developments have accelerated research in vision-based fire and smoke detection systems \citep{frizzi2016convolutional, muhammad2018efficient}.

Despite significant advances in deep learning-based fire detection, most existing approaches focus primarily on classification, detection, or segmentation of fire regions \citep{frizzi2016convolutional, muhammad2018efficient, Gragnaniello2025-jn, Huang2022-po}. While such systems can identify the presence of fire with high accuracy, they often lack contextual awareness regarding the spatial relationship between fire and surrounding objects. 

In safety-critical industrial and engineering environments, risk is not solely determined by the existence of fire but by its proximity to vulnerable assets such as personnel, machinery, fuel sources, or infrastructure. Current detection frameworks rarely incorporate quantitative spatial risk assessment or real-world distance estimation directly within the detection pipeline. Furthermore, pixel-level detections are typically not translated into interpretable physical measurements, limiting their practical utility for operational decision-making. Thus, there remains a gap in the literature for an integrated system that combines high-accuracy fire segmentation with real-time proximity estimation and contextual risk quantification.

To address this gap, this study proposes an enhanced YOLOv8-based dual-model framework for intelligent fire detection and spatial proximity analysis. Beyond accurate fire and smoke segmentation, the framework introduces a quantitative method for estimating real-world distances between detected fire regions and surrounding objects. 

The main contributions of this research are as follows:

\begin{itemize}
    \item Development of a high-performance YOLOv8 instance segmentation model trained on a diverse fire dataset to achieve robust fire and smoke detection across complex environments.
    \item Design of a dual-model architecture that integrates a fire-specific segmentation model with a COCO-pretrained object detection model for contextual scene understanding.
    \item Introduction of a proximity-based risk analysis mechanism that computes pairwise Euclidean distances between fire regions and nearby objects.
    \item Implementation of a pixel-to-meter conversion strategy to translate image-based distances into approximate real-world measurements for interpretable spatial risk estimation.
    \item Provision of annotated visual outputs that transform detection results into actionable spatial intelligence suitable for industrial and safety-critical applications.
\end{itemize}

The remainder of this paper is structured as follows. Section~\ref{sec:related} reviews relevant literature on fire detection and spatial risk analysis. Section~\ref{sec:methods} presents the proposed methodology, including dataset preparation, model architecture, and proximity computation strategy. Section~\ref{sec:results} reports experimental results and performance evaluation. Finally, Section~\ref{sec:conclusion} concludes the paper and outlines future research directions.

\section{Background and Related Work}
\label{sec:related}

Advances in computer vision and machine learning have significantly transformed automated hazard detection systems in recent years. With the widespread deployment of surveillance cameras and improvements in computational hardware, vision-based monitoring systems have become increasingly viable for detecting dangerous events such as fires in real time. Compared with traditional sensor-based detection systems, camera-based approaches offer broader spatial coverage and richer contextual information about the surrounding environment. This enables not only the identification of fire events but also the analysis of their interactions with nearby objects and infrastructure.

Early research in vision-based fire detection focused primarily on handcrafted visual features derived from color, motion, and temporal characteristics of flames. While these approaches established the foundation for automated fire monitoring, they often struggled to generalize under varying illumination conditions, background clutter, and dynamic environments. As a result, researchers increasingly explored machine learning and deep learning techniques to improve detection robustness and adaptability.

The following subsections review the major strands of research relevant to this work, including classical image processing approaches, deep learning-based fire detection methods, and emerging efforts toward spatial awareness and contextual risk analysis in computer vision systems.

\subsection{Image Processing Approaches for Fire Detection}

Prior to the widespread adoption of deep learning models, many fire detection systems were developed using traditional image processing techniques. These methods typically relied on analyzing distinctive visual characteristics of flames, including their color distribution, dynamic motion patterns, and irregular spatial structures. Common techniques included color thresholding in RGB or HSV color spaces, background subtraction, edge detection, and region-based segmentation \citep{celik2007fire}.

Several studies have continued to investigate these image processing approaches in modern fire detection frameworks. For instance, recent work has explored enhanced color and texture-based segmentation techniques to identify flame regions within complex scenes. Other approaches incorporate wavelet transforms, morphological filtering, or feature extraction methods to highlight the characteristic visual signatures of fire prior to classification. These methods demonstrate that carefully designed preprocessing techniques can improve the separability of flame regions from visually similar objects such as streetlights or reflections \citep{Das2025-ip}, solid waste \citep{Nnamoko2022-as}, etc.

Additional studies have proposed hybrid frameworks that combine image processing techniques with machine learning classifiers to improve detection performance. Such approaches typically use image processing pipelines to enhance flame-related visual features before applying classification models to distinguish fire from non-fire patterns. These hybrid systems have shown promising results in scenarios where illumination changes and background interference can significantly affect detection accuracy \citep{Boyd2024-fr, Ali2025-mj, Mhdawi2025-jl}.

Although traditional image processing approaches offer advantages such as lower computational complexity and suitability for embedded systems, they often depend heavily on handcrafted features and environment-specific assumptions. Consequently, their performance may degrade when applied to diverse real-world conditions with varying lighting, occlusion, or background textures. These limitations have motivated the adoption of data-driven deep learning techniques capable of learning more robust feature representations directly from large datasets.

\subsection{Deep Learning for Fire Detection}

The emergence of deep learning has significantly improved the performance of computer vision systems across a wide range of detection and recognition tasks. Convolutional neural networks (CNNs) have demonstrated strong capabilities in learning hierarchical feature representations directly from image data, enabling more robust object detection and classification compared to traditional feature engineering approaches \citep{ren2015faster, redmon2016yolo}.

Two-stage object detection architectures such as Faster R-CNN \citep{ren2015faster} introduced region proposal mechanisms that improved localization accuracy for complex visual objects. However, their relatively high computational cost limits their suitability for real-time surveillance applications where rapid detection is essential.

To address these limitations, single-stage detectors such as YOLO (You Only Look Once) were developed to perform object detection in a single forward pass through the network \citep{redmon2016yolo}. This design enables significantly faster inference while maintaining competitive accuracy. Subsequent improvements to the YOLO architecture have further enhanced detection performance, particularly in real-time monitoring scenarios where computational efficiency is critical \citep{redmon2018yolov3}.

Several studies have applied CNN-based architectures specifically to fire and smoke detection tasks. \cite{frizzi2016convolutional} demonstrated the effectiveness of deep convolutional networks in distinguishing fire regions from visually similar patterns such as artificial lighting or reflections. Similarly, \cite{muhammad2018efficient} proposed an efficient CNN-based framework for early fire detection in surveillance videos, incorporating spatial and temporal analysis to improve detection reliability.

More recent work has explored hybrid approaches that combine deep learning with complementary analytical techniques. For example, \cite{Gragnaniello2025-jn} proposed \textit{FLAME}, a real-time video fire detection system that integrates a deep neural network-based object detector with motion analysis. The motion filtering mechanism reduces false positives by verifying that detected fire regions exhibit characteristic temporal dynamics. Similarly, \cite{Huang2022-po} developed a hybrid architecture combining wavelet-based feature enhancement with CNN classification, improving robustness under complex lighting conditions.

Recent advances in the YOLO family of detectors have further improved real-time detection performance. YOLOv3 introduced multi-scale feature extraction and improved feature pyramid networks to enhance detection of small objects such as distant flames \citep{redmon2018yolov3}. More recently, YOLOv8 has introduced architectural improvements including anchor-free detection, enhanced feature fusion mechanisms, and built-in instance segmentation capabilities \citep{jocher2023yolov8}. These improvements make YOLOv8 particularly well suited for tasks requiring accurate localization and segmentation of fire regions within complex environments.

Despite these advances, most deep learning-based fire detection systems focus primarily on identifying the presence of fire or smoke within images. While such systems achieve high detection accuracy, they often provide limited contextual information about the spatial relationships between detected fire regions and other objects within the scene.

\subsection{Spatial Awareness and Risk Assessment in Vision Systems}

While object detection accuracy has improved substantially in recent years, relatively fewer studies extend detection outputs into spatial reasoning and risk analysis. In many real-world scenarios, the severity of a fire incident depends not only on its presence but also on its proximity to critical assets such as people, machinery, fuel sources, or infrastructure.

Distance estimation from visual data has traditionally been addressed through approaches such as stereo vision, depth sensors, monocular depth estimation networks, and geometric calibration techniques. Monocular image-based approaches are particularly attractive for surveillance environments because they can operate using standard camera infrastructure without requiring specialized depth sensors. However, integrating distance estimation directly into hazard detection pipelines remains relatively underexplored.

In industrial safety contexts, risk assessment frameworks typically combine hazard identification with exposure and vulnerability analysis. Yet, most computer vision-based fire detection systems terminate at the detection stage by providing bounding boxes or segmentation masks without quantifying the spatial threat posed by detected fire regions. Translating pixel-level detections into interpretable real-world distances could significantly enhance the operational value of automated fire monitoring systems.

Recent developments in multi-object detection frameworks trained on large-scale datasets such as COCO enable simultaneous recognition of multiple object categories within complex scenes. Integrating such contextual object detection capabilities with fire detection models offers the potential to analyze interactions between hazards and surrounding entities. However, few existing fire detection systems combine accurate fire segmentation with contextual object detection and real-world proximity estimation within a unified framework.

\subsection{Research Gap}

The existing body of literature demonstrates strong progress in improving fire detection accuracy through both image processing techniques and deep learning-based object detection models. In particular, YOLO-based architectures have achieved impressive real-time performance, enabling practical deployment in surveillance and monitoring applications.

However, current fire detection systems largely operate at the level of hazard identification without incorporating contextual spatial reasoning. Most models focus exclusively on detecting fire or smoke regions without analyzing their proximity to surrounding objects or translating pixel-level detections into interpretable real-world measurements.

This limitation reduces the practical utility of automated fire detection systems in safety-critical environments where risk is determined not only by the presence of fire but also by its spatial relationship to vulnerable assets. There remains a lack of integrated frameworks that combine high-accuracy fire segmentation, contextual object detection, and quantitative spatial risk estimation.

To address this gap, the present study proposes a dual-model framework that integrates YOLOv8-based fire segmentation with contextual object detection and a pixel-to-meter conversion mechanism for estimating real-world distances. By incorporating proximity-based risk analysis directly within the detection pipeline, the proposed approach aims to transform conventional fire detection into a spatially aware hazard monitoring system suitable for industrial and engineering applications.

\section{Materials \& Methods}\label{sec:methods}

This section describes the dataset, model architectures, training configuration, and the end-to-end inference pipeline used to achieve fire/smoke instance segmentation, contextual scene understanding, proximity estimation in meters, and the downstream proximity-aware risk scoring presented in this study. :contentReference[oaicite:0]{index=0} :contentReference[oaicite:1]{index=1}

\subsection{Dataset and annotation}
A diverse fire dataset comprising 9{,}860 images was used to train and evaluate the instance segmentation model. The dataset was pre-annotated for segmentation tasks to enable pixel-level delineation of fire and smoke regions across complex environments. The data included varied visual contexts such as residential scenes, outdoor forested areas, and industrial environments, which supports generalization under different lighting conditions, backgrounds, and occlusions.

A train/validation split was applied, where 78\% of the images (7{,}710) were allocated for training and 22\% (2{,}150) were reserved for validation to assess generalization on unseen data. 

\subsection{Preprocessing and data augmentation}
All images were resized to a fixed resolution of $640 \times 640$ pixels to standardize input dimensions and optimize training efficiency. Standard augmentations were applied during training to improve robustness, including random flipping, scaling, and mosaic augmentation. Augmentation and data loading were handled through the YOLOv8 training pipeline. 

\subsection{Model architecture}
A dual-model architecture was implemented to provide both fire localization and contextual scene understanding. The primary model is a custom-trained YOLOv8 instance segmentation network specialized for fire and smoke segmentation. A secondary YOLOv8 detector pretrained on the COCO dataset was used for general object detection to identify surrounding entities such as people, vehicles, and infrastructure components. Running both models sequentially on each frame enables the system to infer spatial relationships between detected fire regions and nearby objects of interest. 

\subsection{Training configuration and compute environment}
Model training was conducted in the Google Colab environment using an NVIDIA L4 GPU to accelerate optimization and enable efficient experimentation within session constraints. The YOLOv8 segmentation model was trained using a batch size of 12. The study reports extensive training (up to 200 epochs) to obtain stable convergence and strong mask-level performance, and training dynamics were monitored via loss curves and validation metrics. 

\subsection{Inference pipeline for detection and proximity estimation}
For each input (static images or image sequences such as \texttt{.png}, \texttt{.jpg}, and animated \texttt{.gif}), the fire/smoke segmentation model is executed first to produce fire instance masks and bounding boxes. The COCO-pretrained detector is then executed on the same frame to produce surrounding object bounding boxes and class labels. Bounding box centroids are computed for fire instances and surrounding objects, and pairwise Euclidean distances in pixel coordinates are calculated between fire and object centroids to quantify proximity. 

To improve interpretability, pixel distances are converted into approximate real-world distances (meters) using a pixels-per-meter scaling factor derived from known object dimensions or scene calibration. Distances are visualized by overlaying horizontal red proximity lines between fire and nearby objects and printing the corresponding distance values in meters directly on the frame. Annotated frames can be saved individually or compiled into an animated sequence to preserve temporal context for rapid review and post-event analysis. 

\subsection{Distance interpretation and practical limitations}
The pixel-to-meter conversion assumes an estimated reference scale and therefore the metric distances should be interpreted as approximations. The accuracy of distance estimates may vary with camera angle, lens distortion, image resolution, and depth variation because 2D bounding boxes represent monocular projections and may not capture true 3D separation. In the absence of camera calibration or reliable in-frame physical references, the reported distances provide scalable and practical proximity indicators for risk-oriented monitoring rather than exact metrology. 

\subsection{Integration with proximity-aware risk assessment}
The proximity outputs from the dual-model pipeline (detected fire instances, surrounding object categories, and estimated metric distances) are used as direct inputs to the proximity-aware risk assessment model introduced in this paper. In this integration, risk is determined by combining fire evidence (segmentation confidence and spatial extent) with object relevance (class-driven vulnerability) and distance-driven exposure decay. This transforms detection outputs into quantitative hazard prioritization suitable for automated alerting and decision support in engineering and industrial contexts. 

\section{Experimental Results}\label{sec:results}
This study propose an enhanced YOLOv8 segmentation architecture to develop a deep learning model capable of detecting fire  in various visual environments in regard to spatial relationship with other objects and proximity calculations. The dataset used consisted of 9,860 images that were pre-annotated for object segmentation tasks. Beyond detection, a key contribution of this work is the implementation of a proximity-based risk analysis framework. By using a dual-model setup—one trained to detect fire and another pretrained on general objects—the system not only identifies fire presence but also calculates the spatial distance between fire regions and nearby objects of interest. These distances are converted from pixels to real-world meters using a calibrated scaling factor, allowing for quantitative assessment of fire risk in context. The annotated outputs visualize fire locations, object categories, and distances through overlaid labels and red proximity lines, and are compiled into an animated .gif for intuitive analysis. This approach enables enhanced decision-making for engineering and industrial environments by transforming raw detection data into actionable spatial intelligence. A dataset split was applied in which 78\% of the images, equivalent to 7,710, were allocated to the training set, while the remaining 22\%, comprising 2,150 images, were assigned to the validation set. This split ensured that the model had sufficient exposure to learn from diverse fire-related scenarios while also retaining a significant portion of data for evaluating generalization performance. The training process was conducted on the Google Colab platform, utilizing an NVIDIA L4 GPU unit. This allowed for efficient training of the YOLOv8 model, benefiting from the GPU's high processing power and memory optimization features, which are well-suited for deep learning tasks such as object detection and segmentation. The input images were resized to 640 by 640 pixels to standardize the dataset and optimize training performance. The model was trained over 100 epochs using a batch size of 12 and incorporated standard image augmentations including random flipping, scaling, and mosaic techniques to enhance robustness. The YOLOv8 framework automatically handled augmentation and data loading within each training iteration.

During training, the model demonstrated consistent improvements in loss minimization and detection accuracy. The training and validation loss curves showed a stable learning trajectory, with validation performance peaking around epoch 12. At this point, the model achieved its best mAP@0.5, indicating a high degree of accuracy in identifying and segmenting fire and smoke regions within unseen images. Visual inspection of the validation outputs confirmed that the model was able to delineate fire and smoke contours with considerable precision, even in challenging conditions such as low lighting or background clutter. Moreover, the effectiveness of the model can be attributed to the diversity and quality of the dataset, which featured fire occurrences in multiple settings including residential areas, outdoor forested regions, and industrial environments. The model’s ability to accurately predict masks for both fire and smoke suggests strong potential for real-world applications in surveillance, safety monitoring, and emergency response systems. Nonetheless, the use of the L4 GPU enabled faster convergence times and allowed for the exploration of different configurations without exceeding session limitations on Google Colab. Overall, the YOLOv8-based model trained under this configuration proved to be efficient and reliable, with results that validate its suitability for high-accuracy fire detection and segmentation tasks.

\begin{figure}[tbh!]
\centering

\begin{subfigure}[b]{0.48\textwidth}
  \centering
  \includegraphics[width=\textwidth]{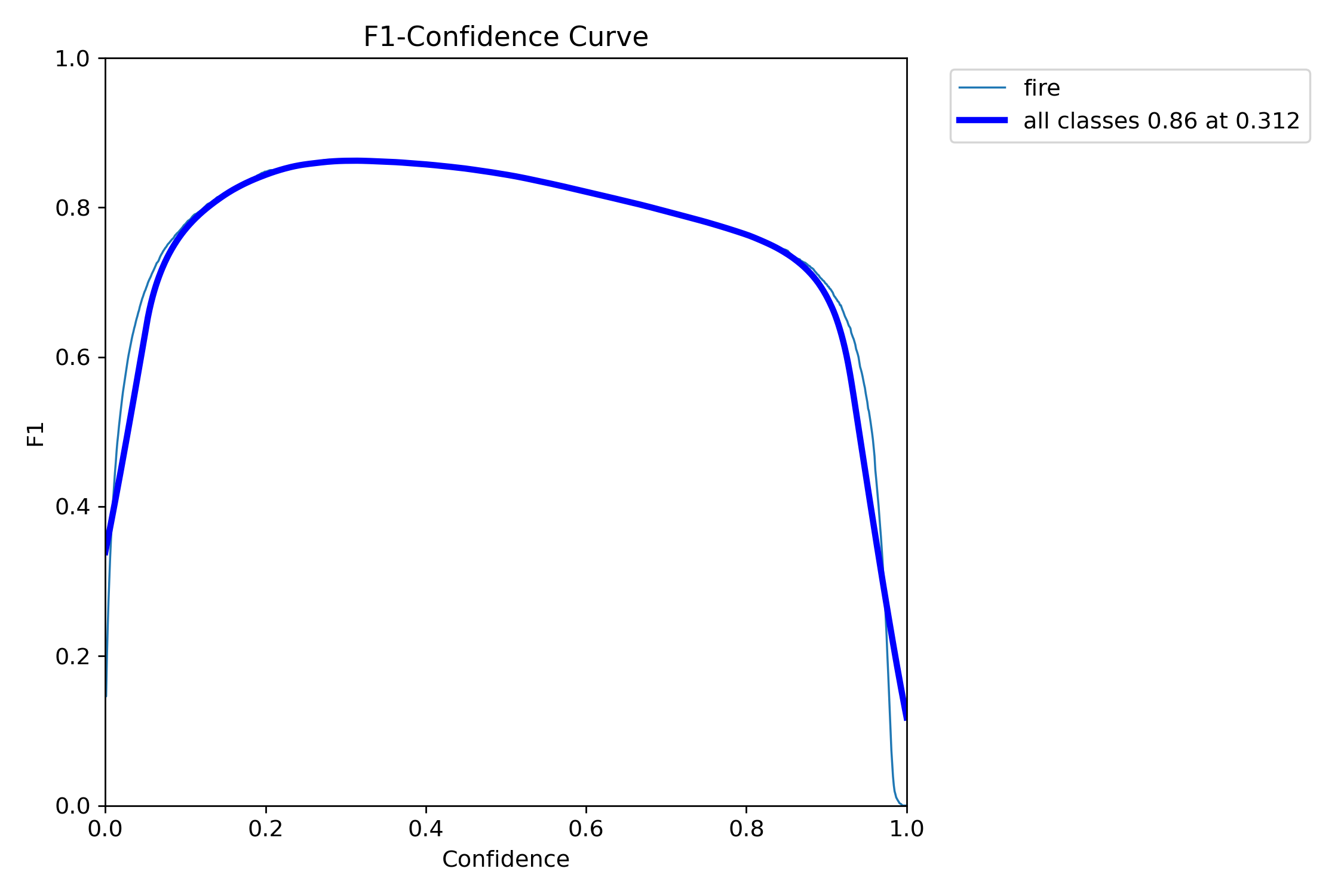}
  \caption{F1 score versus confidence threshold.}
  \label{fig:f1curve1}
\end{subfigure}
\hfill
\begin{subfigure}[b]{0.48\textwidth}
  \centering
  \includegraphics[width=\textwidth]{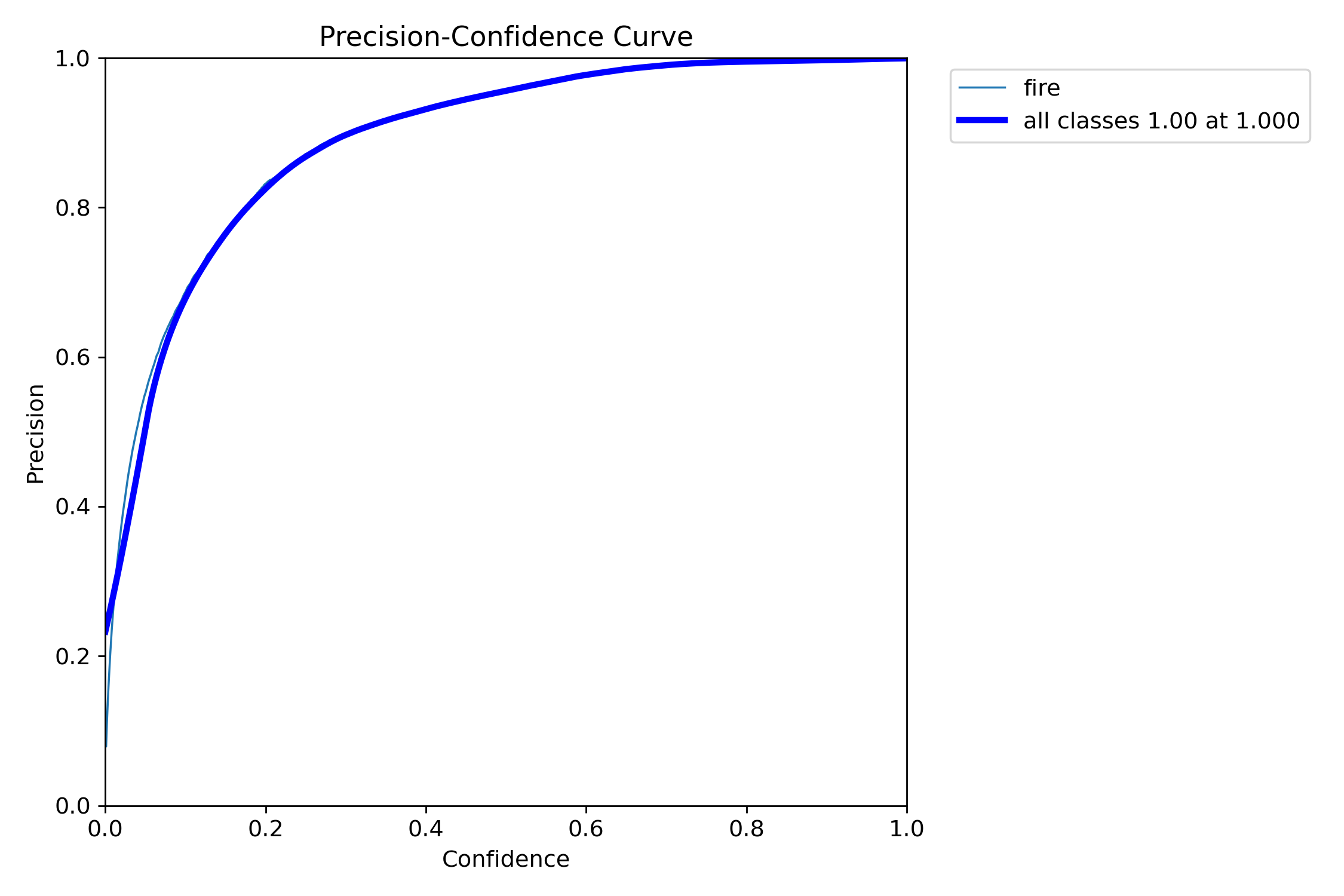}
  \caption{Precision versus confidence threshold.}
  \label{fig:precisioncurve}
\end{subfigure}

\caption{Model performance as a function of confidence thresholds.}
\label{fig:perf_confidence}
\end{figure}

\noindent
In Figure \ref{fig:f1curve1} shows the F1-confidence curve, which evaluates the harmonic mean of precision and recall at different confidence thresholds. The peak F1 score of 0.86 occurs at a threshold of 0.312, indicating a well-balanced model performance when both false positives and false negatives are minimized.

\noindent
While Figure \ref{fig:precisioncurve} presents the precision-confidence curve. This visualisation confirms that the model achieves perfect precision (1.0) at a confidence threshold of 1.0, meaning that all detections at that level are correct. However, higher precision at extreme thresholds often comes with a reduction in recall, limiting the number of predictions made.

\begin{figure}[tbh!]
\centering

\begin{subfigure}[b]{0.48\textwidth}
  \centering
  \includegraphics[width=\textwidth]{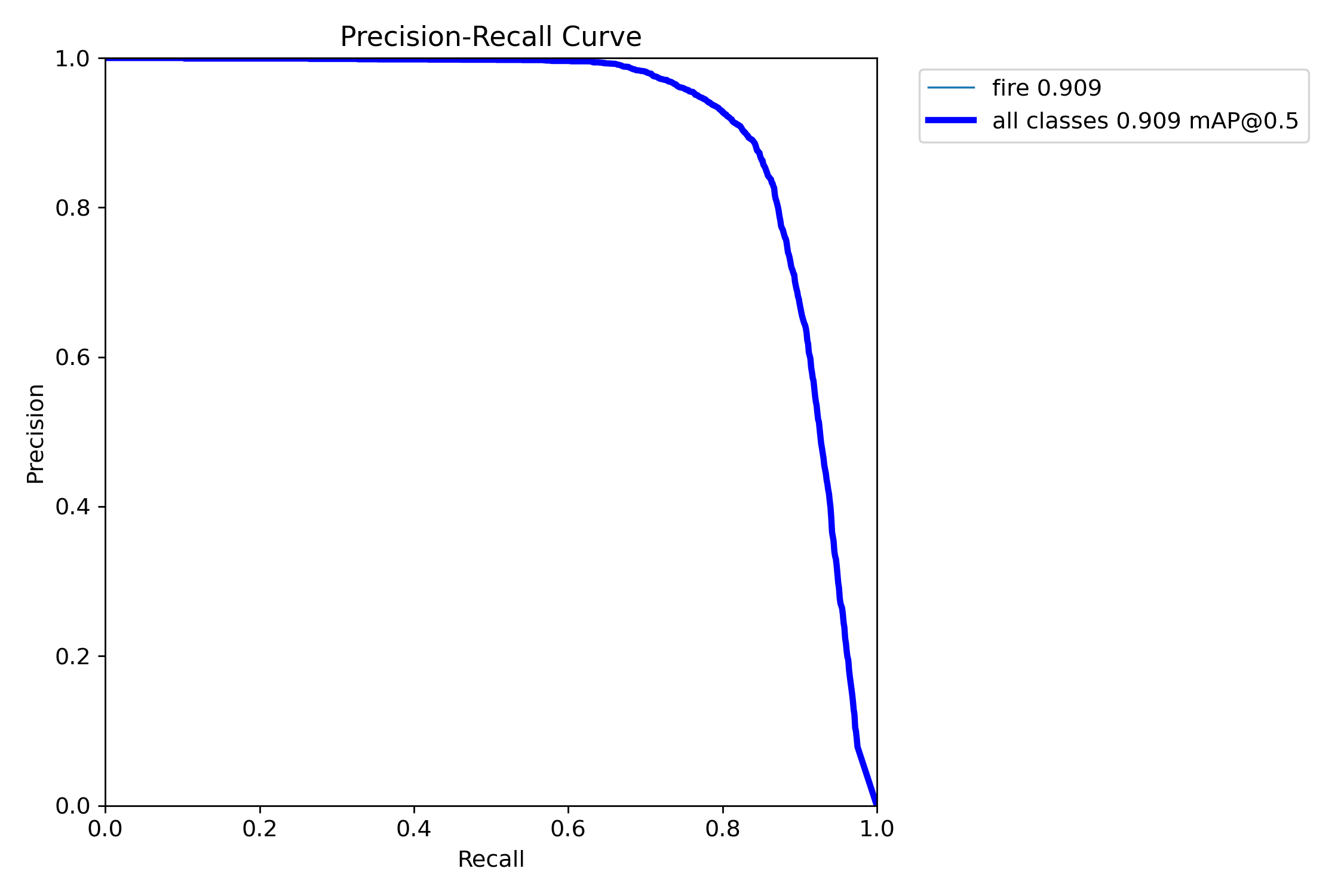}
  \caption{Precision versus recall curve.}
  \label{fig:prcurve}
\end{subfigure}
\hfill
\begin{subfigure}[b]{0.48\textwidth}
  \centering
  \includegraphics[width=\textwidth]{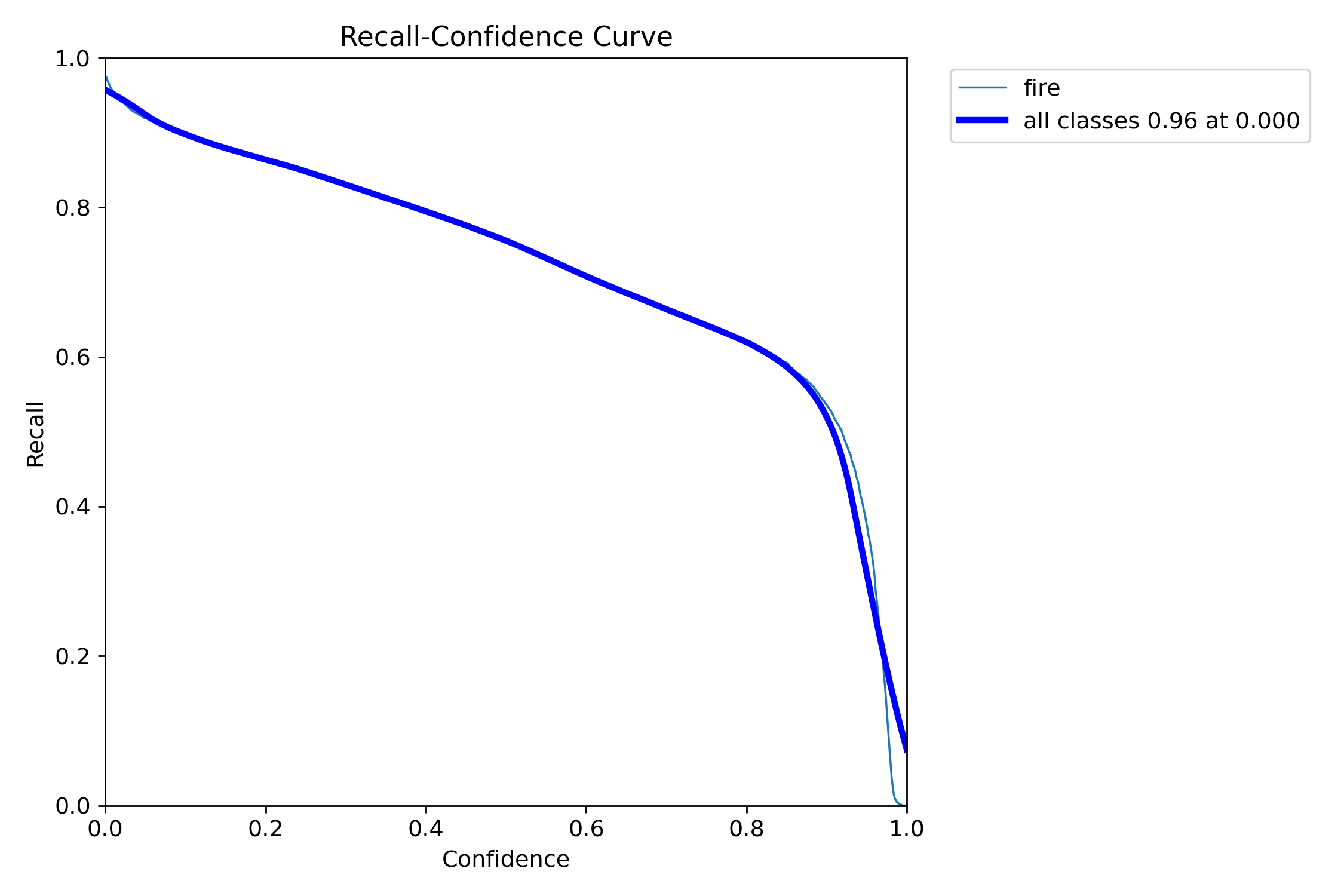}
  \caption{Recall versus confidence threshold.}
  \label{fig:recallcurve}
\end{subfigure}

\caption{Evaluation curves for model performance.}
\label{fig:evaluation_curves}
\end{figure}

\noindent
Additionally, Figure \ref{fig:prcurve} depicts the precision-recall curve, where the model achieves a mean average precision (mAP@0.5) of 0.909. This reflects excellent performance in maintaining high recall and precision across varying prediction thresholds, affirming the reliability of the detection system under realistic conditions.

\noindent
Moreover, Figure \ref{fig:recallcurve} illustrates the recall-confidence curve, where the maximum recall of 0.96 is achieved at a very low threshold. This is expected, as lower thresholds allow the model to classify more instances as positive (fire), which maximizes sensitivity but may introduce false positives.

\begin{figure}[tbh!]
\centering

\begin{subfigure}[b]{0.48\textwidth}
  \centering
  \includegraphics[width=\textwidth]{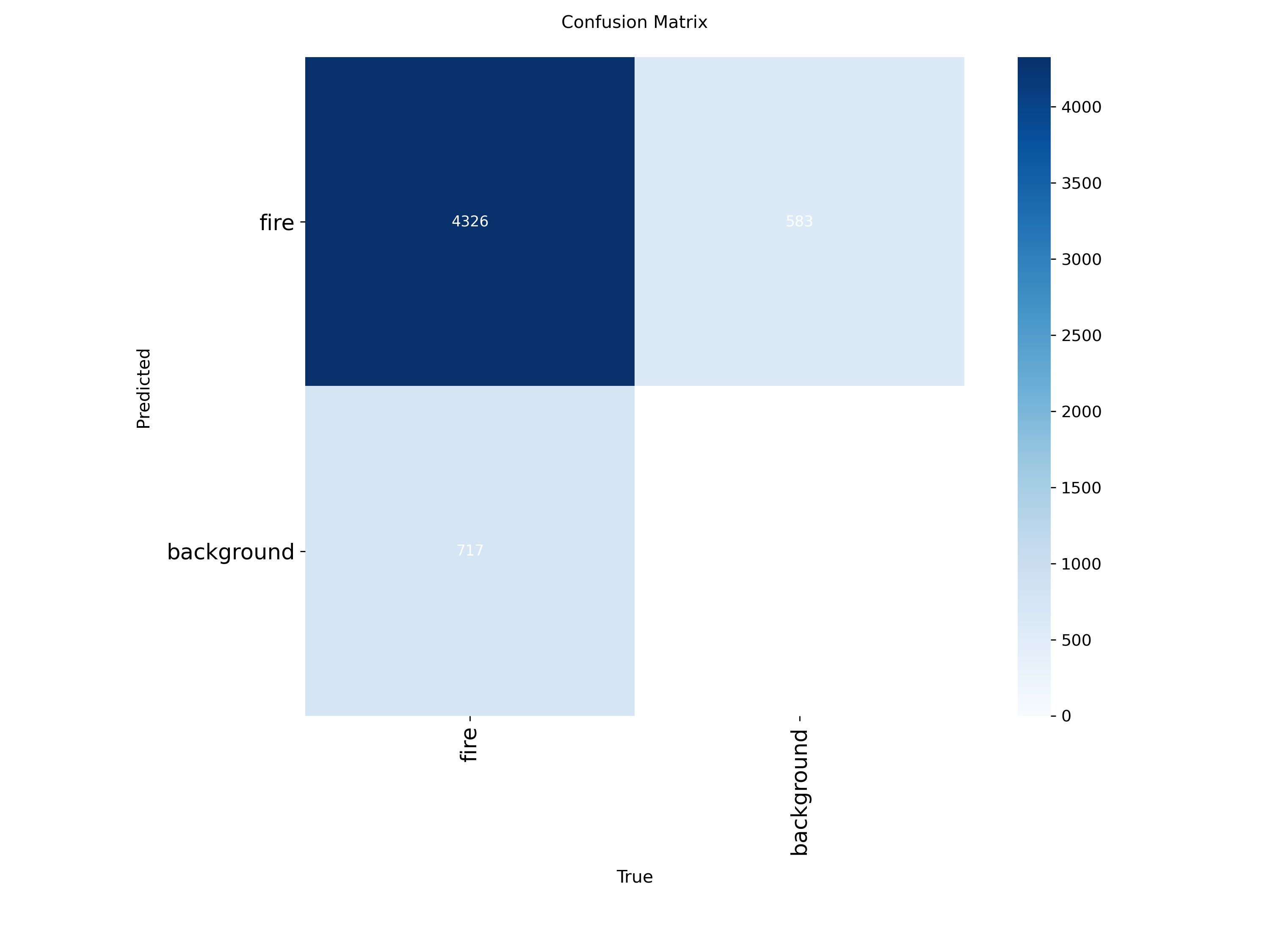}
  \caption{Confusion matrix (raw counts).}
  \label{fig:confusionmatrix}
\end{subfigure}
\hfill
\begin{subfigure}[b]{0.48\textwidth}
  \centering
  \includegraphics[width=\textwidth]{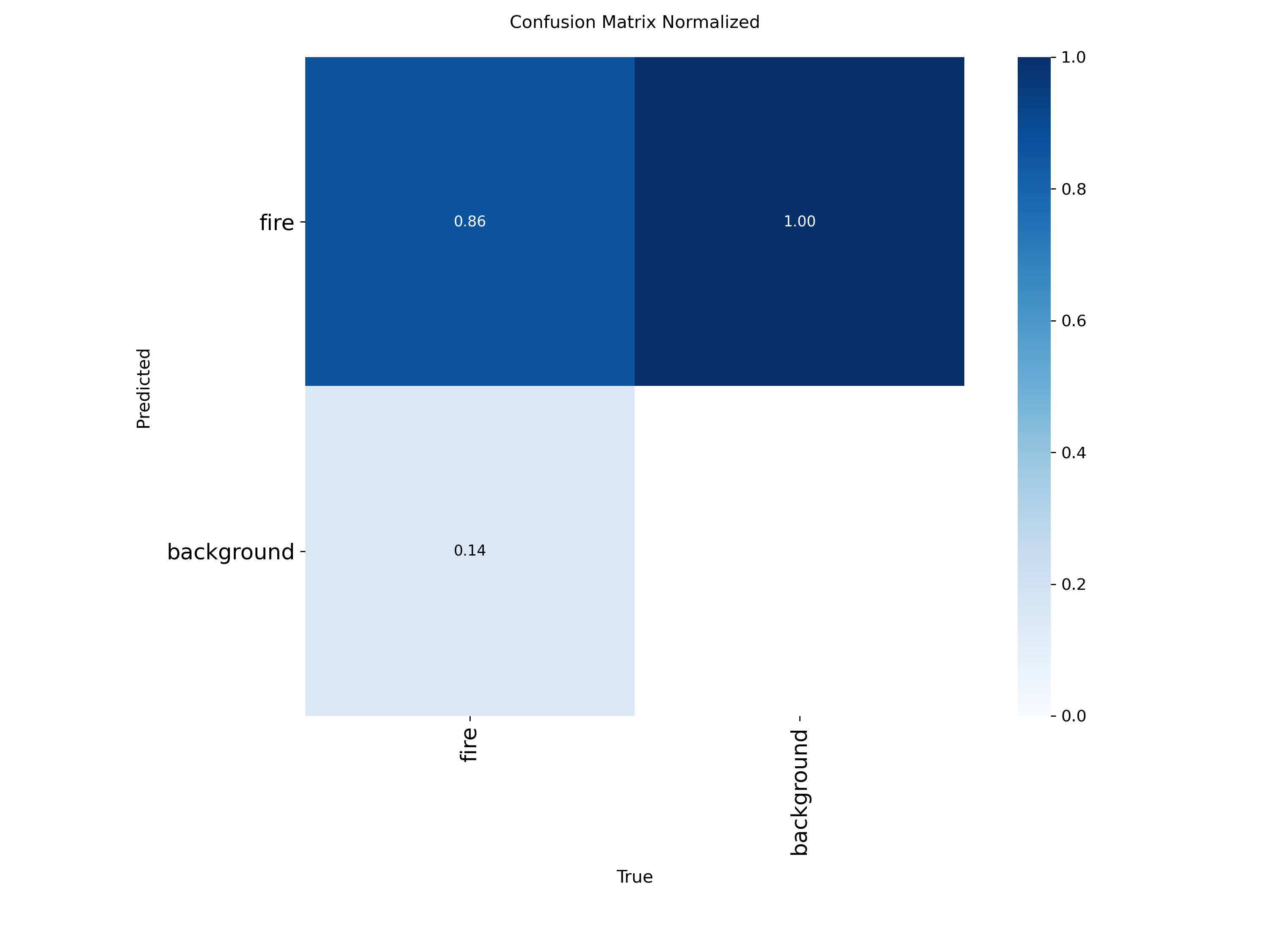}
  \caption{Normalized confusion matrix.}
  \label{fig:normconfmatrix}
\end{subfigure}

\caption{Comparison of raw and normalized confusion matrices for fire detection performance.}
\label{fig:confmatrix_combined}
\end{figure}

\noindent
Nonetheless, Figure \ref{fig:confusionmatrix} displays the raw confusion matrix of prediction outcomes. The model correctly predicted 4,326 fire instances, with 717 missed (false negatives) and 583 incorrect fire predictions (false positives). This result indicates strong but improvable classification accuracy, particularly for reducing missed detections in safety-critical applications.

\noindent
It can be noticed in Figure \ref{fig:normconfmatrix} shows the normalized confusion matrix. Here, 86\% of true fire instances are correctly classified, and 14\% are misclassified as background. These values validate the model’s bias toward sensitivity, which is beneficial in early fire detection scenarios where false negatives are more costly than false positives.

\begin{figure}[tbh!]
\centering

\begin{subfigure}[b]{0.45\textwidth}
  \centering
  \includegraphics[width=\textwidth]{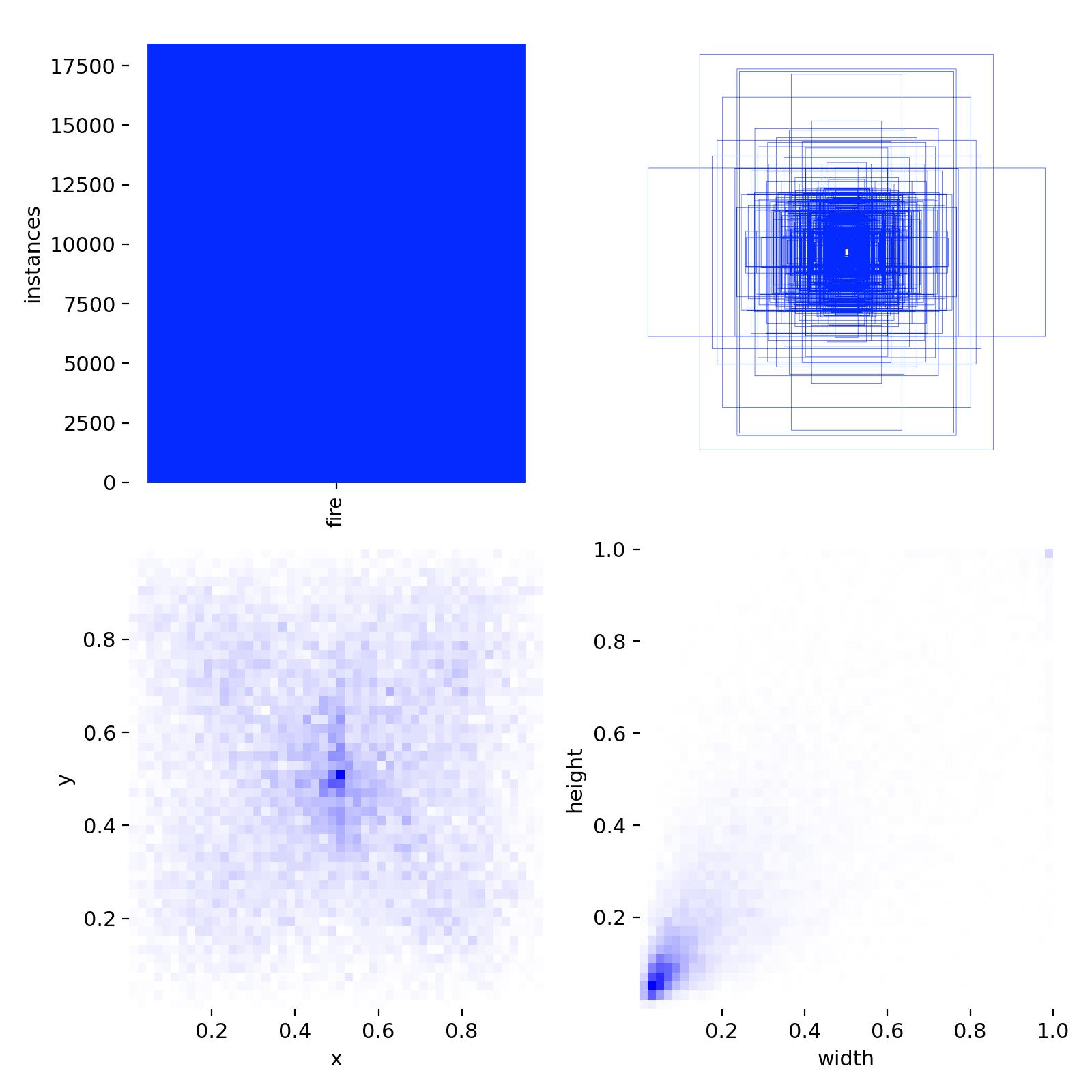}
  \caption{Instance and anchor box heatmaps.}
  \label{fig:anchorheatmap}
\end{subfigure}
\hfill
\begin{subfigure}[b]{0.5\textwidth}
  \centering
  \includegraphics[width=\textwidth]{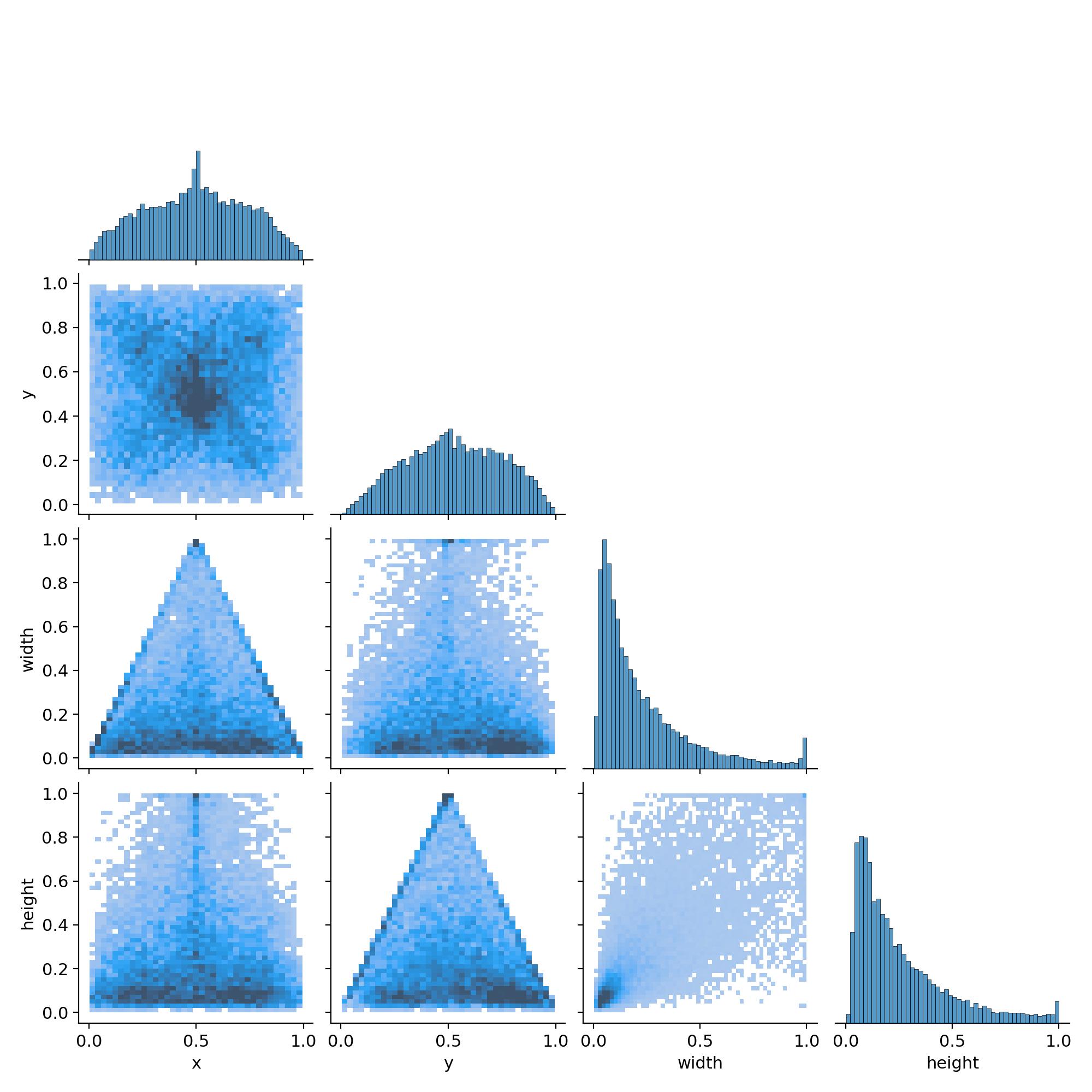}
  \caption{Pairplot of object detection annotations.}
  \label{fig:pairplot}
\end{subfigure}

\caption{Visualizations of object detection spatial analysis.}
\label{fig:object_vis_analysis}
\end{figure}

\noindent
Moreover, Figure \ref{fig:anchorheatmap} provides a visual summary of bounding box annotations, showing the frequency of fire instances, the shape distribution of bounding boxes, and their relative positions in the images. Most bounding boxes are concentrated around the image center with consistent width and height ratios, which supports model stability in anchor selection and feature learning. \noindent
Furthermore, Figure \ref{fig:pairplot} presents a pairplot of annotation metadata, including x and y positions, width, and height of bounding boxes. The strong clustering around lower width and height values reveals that most fire annotations cover a small spatial area, and the centralized x-y density supports the model’s robustness when focusing on high-risk image zones. Furthermore, Figure \ref{fig:pairplot} presents a pairplot of annotation metadata, including x and y positions, width, and height of bounding boxes. The strong clustering around lower width and height values reveals that most fire annotations cover a small spatial area, and the centralized x-y density supports the model’s robustness when focusing on high-risk image zones.


\noindent
Additionally, Figure \ref{fig:f1curve2} shows the F1-confidence curve from a second independent training run. This curve nearly replicates the initial results, with a peak F1 score of 0.86 at a threshold of 0.316. The consistency between experiments demonstrates strong model reliability across multiple runs.


\begin{figure}[tbh!]
    \centering
    \begin{subfigure}{0.49\textwidth}
        \centering
        \includegraphics[width=\linewidth]{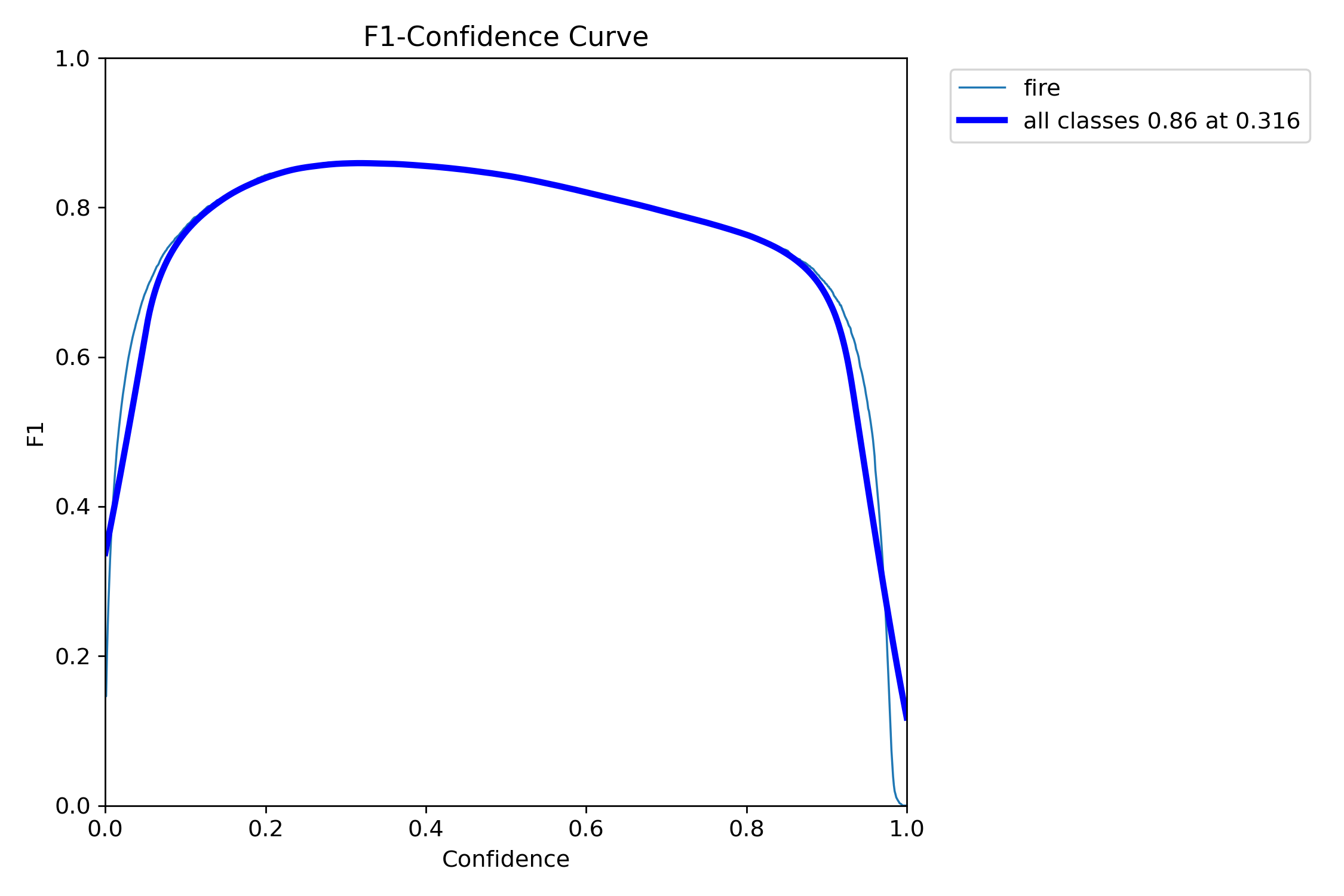}
        \caption{F1-confidence curve from a second training run.}
        \label{fig:f1curve2}
    \end{subfigure}
    \hfill
    \begin{subfigure}{0.49\textwidth}
        \centering
        \includegraphics[width=\linewidth]{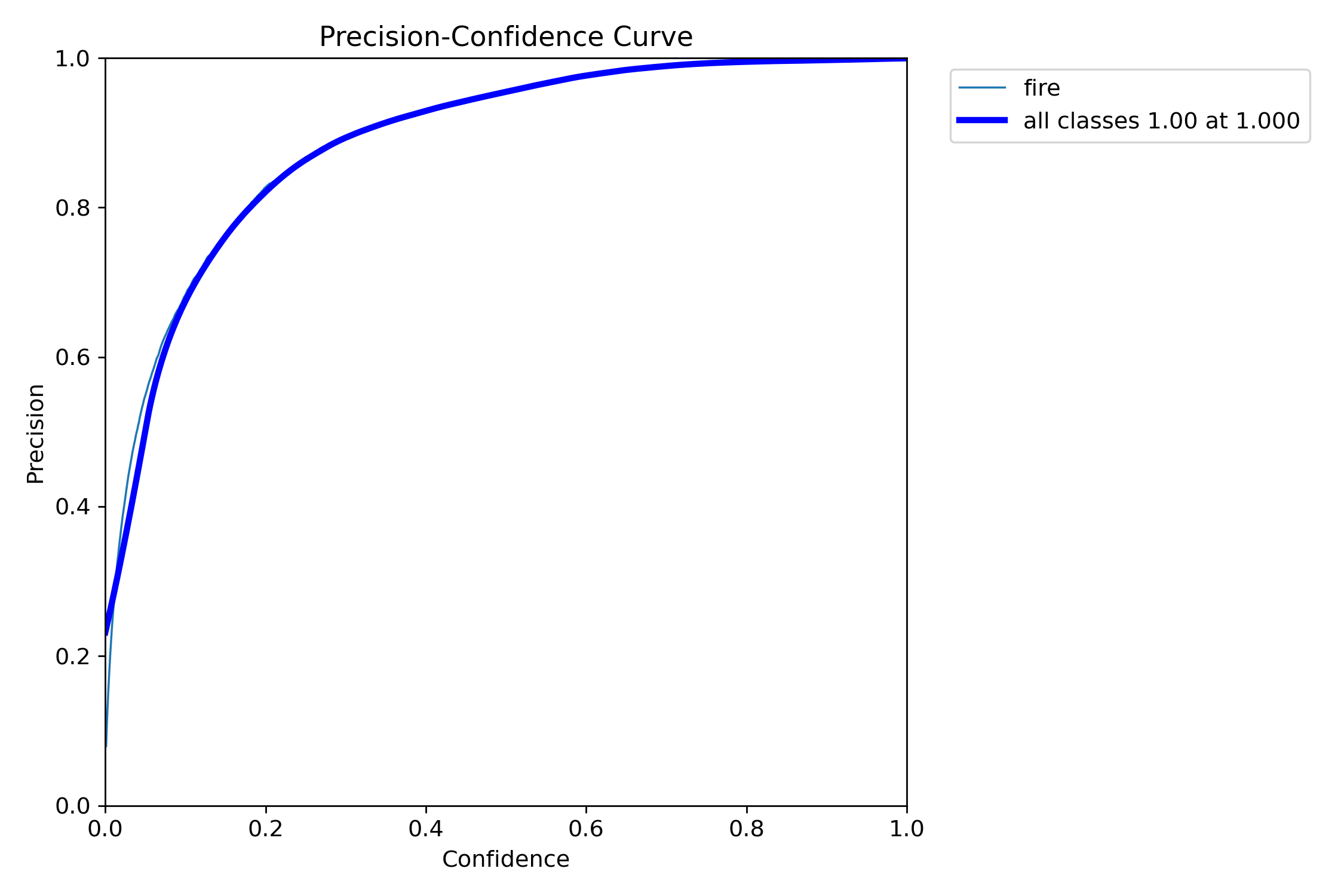}
        \caption{Precision-confidence curve from a second run.}
        \label{fig:precisioncompare}
    \end{subfigure}
    \caption{Precision \& F1-confidence curve from a second training run.}
    \label{fig:f1Andpreccurve2}
\end{figure}

\noindent
While in Figure \ref{fig:precisioncompare} shows the repeated precision-confidence relationship from the comparison run. Again, the model achieves perfect precision at a confidence threshold of 1.0, confirming high confidence predictions are consistently accurate. 
Moreover, Figure \ref{fig:prcurve_updated} presents the precision-recall curve generated from the most recent training session. The model maintains a strong trade-off between precision and recall, culminating in a mean average precision (mAP@0.5) of 0.905. This consistency in performance confirms the model’s stability and ability to detect fire instances reliably across a variety of validation images.



\begin{figure}[tbh!]
    \centering
    \begin{subfigure}{0.49\textwidth}
        \centering
        \includegraphics[width=\linewidth]{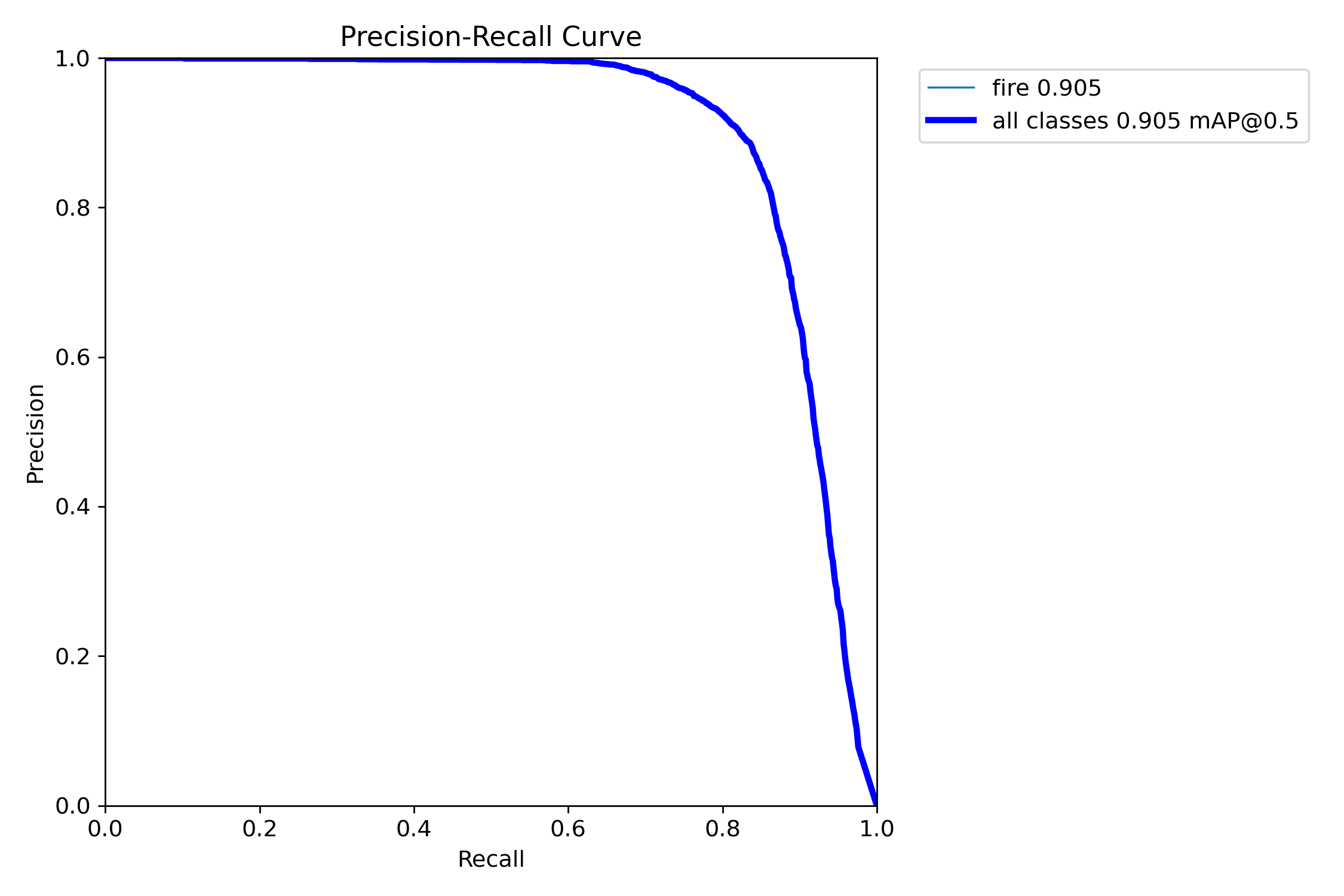}
        \caption{Precision-recall curve (updated model run).}
        \label{fig:prcurve_updated}
    \end{subfigure}
    \hfill
    \begin{subfigure}{0.49\textwidth}
        \centering
        \includegraphics[width=\linewidth]{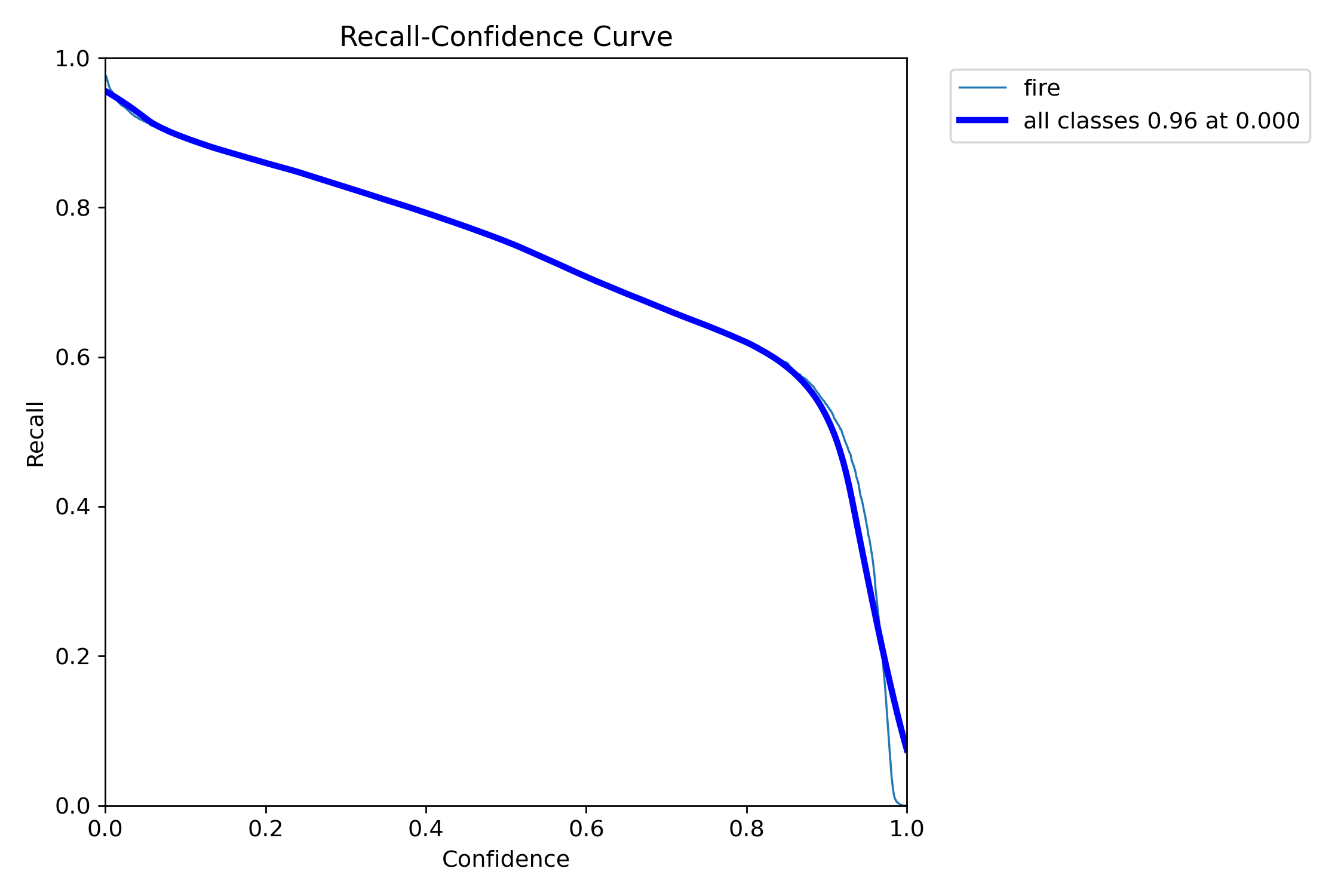}
        \caption{Recall-confidence curve (updated model run).}
        \label{fig:recallcurve_updated}
    \end{subfigure}
    \caption{Precision \& Recall-confidence curve (updated model run).}
    \label{fig:preAndrecallcurve_updated}
\end{figure}

\noindent
Consequently, Figure \ref{fig:recallcurve_updated} shows the recall as a function of confidence threshold. The highest recall of 0.96 is achieved at a threshold of 0.000, reinforcing the expected behavior where low thresholds capture most true positives but may include more false alarms. This metric supports the application of the model in real-time or early-warning systems, where maximizing sensitivity is often more critical than precision.

\begin{figure}[tbh!]
\centering
\includegraphics[width=0.98\textwidth]{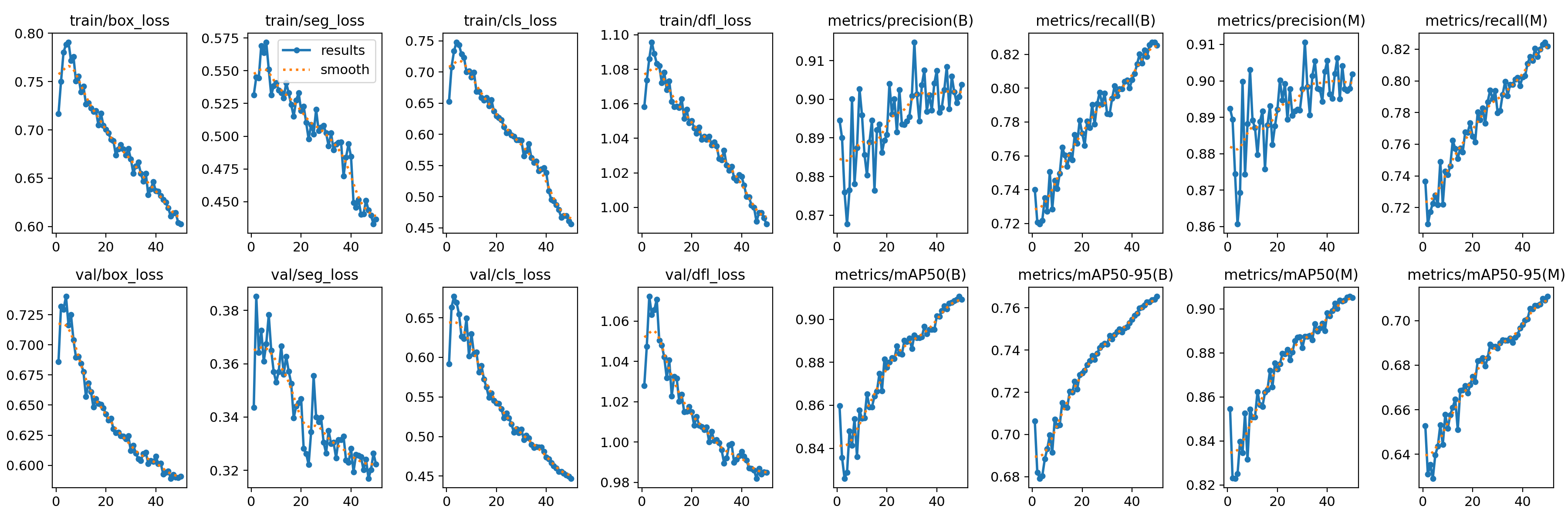}
\caption{Loss curves and metric performance across training epochs.}
\label{fig:lossmetrics}
\end{figure}

\noindent
The overall results of the model training and validation in showin in Figure \ref{fig:lossmetrics} which  illustrates a comprehensive set of evaluation metrics captured during the model's 50-epoch training cycle. These include four primary types of loss functions: bounding box regression loss (`$train:box loss$` and `$val:box loss$`), segmentation loss (`$train:seg loss$` and `$val:seg loss$`), classification loss (`$train:cls loss$` and `$val:cls loss$`), and distributional focal loss (`$train:dfl loss$` and `$val:dfl loss$`). Each loss curve shows a steady and consistent decline, indicating that the model successfully learned to generalize from the training data to the validation set without signs of overfitting. In the precision and recall plots labeled `metrics/precision(B)` and `metrics/recall(B)` for the bounding box predictions, we observe a notable increase over time, reaching approximately 91.2\% precision and 82.5\% recall by the final epoch. These values show that the model correctly predicts most fire instances with minimal false positives and a high degree of coverage. The same trend is seen in the segmentation-based metrics `metrics/precision(M)` and `metrics/recall(M)`, which also reach around 91\% precision and 82.4\% recall, confirming that both bounding box and segmentation predictions are performing with nearly equivalent reliability and accuracy. Moreover, the metrics `metrics/mAP50(B)` and `metrics/mAP50(M)` reveal a peak mean average precision at IoU threshold 0.5 of about 90.4\%, while `metrics/mAP50-95(B)` and `metrics/mAP50-95(M)` show progressively rising curves that settle at around 76.5\% and 70.5\% respectively. These latter metrics provide a more stringent measure of model performance across multiple IoU thresholds, confirming strong spatial precision in both detection and segmentation tasks. Nonetheless, these results reflect a well-trained YOLOv8 segmentation model, with minimal loss values and high performance across all standard object detection benchmarks. This level of accuracy and precision indicates the model is well-suited for practical deployment in fire detection systems where both detection confidence and mask-level precision are critical.

\begin{figure}[h]
\centering
\includegraphics[width=\textwidth]{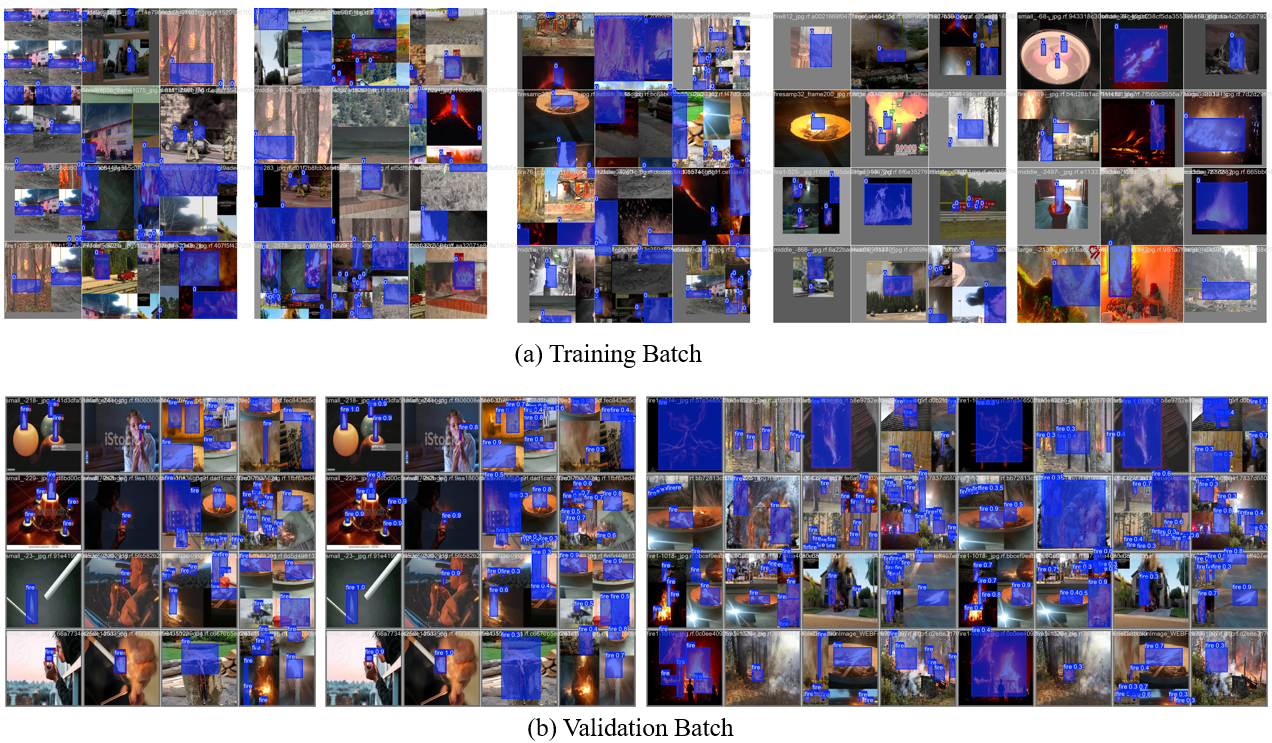}
\caption{Visualisation of predicted masks and bounding boxes on sample training and validation batches.}
\label{fig:trainval_batches}
\end{figure}

\noindent
Moreover, Figure \ref{fig:trainval_batches} illustrates example outputs from both the training and validation batches of the YOLOv8 fire detection model after 200 training epochs. In part (a), we observe a series of training images annotated with blue segmentation masks and bounding boxes, all of which are associated with detected fire instances. These visual samples reflect the model's exposure to a diverse range of fire appearances, environments, lighting conditions, and occlusions.
 Moreover, the predictions in the training batch show high confidence levels, with most bounding boxes labeled accurately and masks closely following the fire contours. The examples include open flames, smoke emissions, and thermal features, demonstrating that the model has generalized well across different visual contexts. While in part (b), the validation batch predictions show equally strong performance. The predicted bounding boxes and segmentation masks cover the true fire regions effectively, even in complex scenes that include background clutter or overlapping fire sources. The confidence scores shown for each prediction range predominantly between 0.7 and 1.0, suggesting high certainty and low noise in the output. Importantly, these visual results further confirm the consistency and reliability of the model following the complete 200-epoch training cycle. The use of extensive augmentation and carefully curated data has contributed to the model’s ability to distinguish between active fire, light reflection, and surrounding objects which is essential in real-world fire monitoring systems.\\
In addition to accurate fire detection using a custom-trained YOLOv8 instance segmentation model, this work introduces a proximity analysis technique aimed at evaluating spatial risk by calculating the distance between detected fire instances and nearby objects of interest within visual surveillance frames. This approach extends traditional detection systems by not only identifying the presence of fire but also providing quantitative assessments of risk based on spatial proximity — a critical factor in engineering and industrial environments.
Our key contribution lies in quantifying the spatial relationship between fire and surrounding objects in real-time. By measuring the distance between detected fire regions and nearby entities (e.g., people, machinery, infrastructure), we can assess the immediate risk posed by the fire's location. This allows us to distinguish between isolated, low-risk fire instances and those that pose an imminent threat to critical assets or personnel, thereby enabling targeted and timely interventions. The system leverages a dual-model setup. A custom YOLOv8 model trained specifically on fire and smoke segmentation is used to detect fire instances, while a second YOLOv8 model pretrained on the COCO dataset performs general object detection. Both models are run sequentially on each input frame, whether as static images or animated ".gif" or "png" or "jpg" sequences. Bounding box predictions are extracted, and the centroids of fire and other objects are used to compute pairwise Euclidean distances. These distances, calculated in pixels, are then converted to approximate real-world measurements (in meters) using a pixels-per-meter scaling factor, which is derived from known object dimensions or scene calibration. Distances between fire and nearby objects are visually represented by horizontal red lines overlayed on each frame, with corresponding distance values annotated directly on the image. This not only improves interpretability but also facilitates real-time or post-event analysis of fire spread risk. The entire annotated sequence is then saved as an image file, preserving temporal context and enabling fast review. This proximity-based analysis framework adds a new dimension to fire detection by transforming spatial relationships into measurable, actionable insights supporting enhanced risk evaluation, automated alerts, and operational decision-making in industrial and engineering settings.

\begin{figure}[h]
\centering
\includegraphics[width=0.6\textwidth]{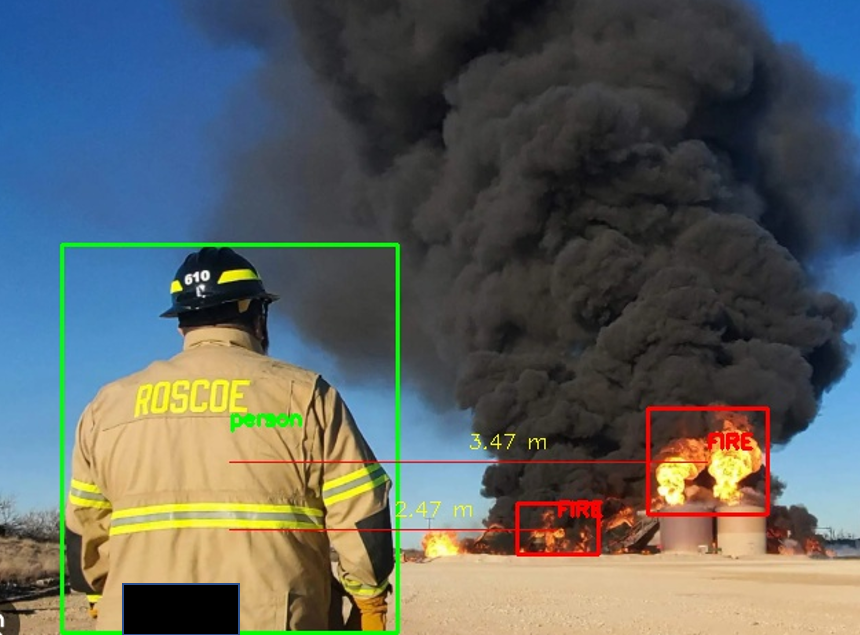}
\caption{Annotated frame showing detected fire regions and surrounding objects with horizontal red lines indicating proximity. Distance values are overlaid in meters using a pixel-to-meter scaling factor and should be treated as approximate unless the camera and scene are calibrated (e.g., known intrinsics/extrinsics and a reliable scale reference). Because the measurements are inferred from 2D image geometry, the estimated distance can be biased by perspective projection (depth variation), camera angle relative to the ground plane, lens distortion, and changes in object size with range; additionally, bounding-box centroids provide only an approximate contact point rather than true nearest-edge separation.}
\label{fig:fire_proximity}
\end{figure}

\noindent
Figure~\ref{fig:fire_proximity} illustrates the output of our proximity-based risk analysis pipeline. The red bounding boxes highlight fire instances, while green boxes denote general objects detected by the COCO-pretrained YOLOv8 model. Red horizontal lines represent the calculated proximity between fire and surrounding objects, with corresponding distances annotated in meters. These measurements are derived using a pixel-to-meter scaling factor.
It is important to note that the accuracy of the distance estimations is subject to several factors. First, the pixel-to-meter conversion assumes a known or estimated reference scale, which may vary depending on camera angle, lens distortion, image resolution, and object depth in the scene. Without camera calibration or real-world object size references within each frame, the estimated distance values should be interpreted as approximations. Additionally, object bounding boxes represent 2D projections that may not account for depth variations, further affecting the real-world spatial accuracy. Despite these constraints, the proximity analysis offers a practical and scalable method for estimating fire risk based on visual cues in uncontrolled environments.

\subsection{Proximity-aware fire hazard risk assessment model}
This work goes beyond fire presence detection by incorporating contextual awareness and spatial reasoning, where risk is interpreted as a function of the detected fire/smoke severity and its proximity to vulnerable surrounding entities. The proposed pipeline follows the dual-model structure adopted in this study: (i) a custom-trained YOLOv8 instance segmentation model for fire and smoke, and (ii) a COCO-pretrained YOLOv8 object detection model for surrounding entities (e.g., people, vehicles, equipment, infrastructure). The outputs of both models are fused per frame to compute pairwise fire-to-object distances using centroid geometry and then convert pixel distances into approximate metric units via a pixels-per-meter scaling factor derived from known object dimensions or scene calibration. This design aligns with the paper’s primary contribution of translating raw visual detections into interpretable, proximity-based spatial risk intelligence suitable for safety-critical settings \cite{zio2025advances} \cite{beljikangarlou2025applying} \cite{kim2026quantitative}. 

\paragraph{Notation and model outputs.}
For each input frame, the fire segmentation model returns a set of detected fire/smoke instances
\begin{equation}
\mathcal{F}=\{f_i\}_{i=1}^{N_F},
\end{equation}
where each instance $f_i$ provides (a) a segmentation mask or polygon region, (b) a bounding box $B^F_i=(x^F_i,y^F_i,w^F_i,h^F_i)$, and (c) a confidence score $s^F_i\in[0,1]$. In parallel, the contextual object detector returns
\begin{equation}
\mathcal{O}=\{o_j\}_{j=1}^{N_O},
\end{equation}
where each object $o_j$ provides a bounding box $B^O_j=(x^O_j,y^O_j,w^O_j,h^O_j)$, a class label $c_j$ (COCO category), and a confidence score $s^O_j\in[0,1]$. These two outputs are computed sequentially for each frame to enable contextual proximity evaluation between fire instances and nearby entities. :contentReference[oaicite:3]{index=3} :contentReference[oaicite:4]{index=4}

\paragraph{Centroid extraction and pixel-distance computation.}
Following the proximity analysis described in this study, the geometric center (centroid) of each fire instance and each detected object is extracted from its bounding box:
\begin{equation}
\mathbf{p}^F_i=\left(x^F_i+\frac{w^F_i}{2},\,y^F_i+\frac{h^F_i}{2}\right), \qquad
\mathbf{p}^O_j=\left(x^O_j+\frac{w^O_j}{2},\,y^O_j+\frac{h^O_j}{2}\right).
\end{equation}
For each fire instance $f_i$, the pixel-space Euclidean distance to object $o_j$ is then computed as
\begin{equation}
d^{px}_{ij}=\|\mathbf{p}^F_i-\mathbf{p}^O_j\|_2=\sqrt{(p^{F}_{i,x}-p^{O}_{j,x})^2+(p^{F}_{i,y}-p^{O}_{j,y})^2}.
\end{equation}
This is consistent with the paper’s use of centroid-based pairwise Euclidean distances for proximity quantification between detected fire regions and surrounding objects of interest. :contentReference[oaicite:5]{index=5} :contentReference[oaicite:6]{index=6}

\paragraph{Pixel-to-meter conversion.}
To produce interpretable distances, the pixel distance is converted into an approximate real-world distance (meters) using a pixels-per-meter scaling factor $\kappa$:
\begin{equation}
d^{m}_{ij}=\frac{d^{px}_{ij}}{\kappa}.
\end{equation}
The scale $\kappa$ is derived using either (i) known dimensions of a reference object present in the scene or (ii) scene calibration. If a reference object of known physical width $W^{m}_{ref}$ corresponds to a measured pixel width $W^{px}_{ref}$ in the same frame (or calibrated view), then
\begin{equation}
\kappa=\frac{W^{px}_{ref}}{W^{m}_{ref}} \quad [\text{px/m}].
\end{equation}
As highlighted in the paper, the resulting metric distances should be interpreted as approximations, since accuracy can be affected by camera angle, lens distortion, image resolution, and depth variation in a monocular 2D projection without full camera calibration. 
\paragraph{Risk model structure.}
In industrial safety contexts, risk is not determined solely by the presence of fire but by its proximity to vulnerable assets such as personnel, machinery, fuel sources, or infrastructure. Accordingly, we define a proximity-aware risk score that combines (i) fire detection strength and extent, (ii) object vulnerability (by class), and (iii) spatial exposure via distance. This operationalizes the paper’s motivation to bridge detection outputs with actionable spatial risk intelligence by translating pixel-level detections into interpretable physical measurements for decision-making \cite{wei2026fire} \cite{morchid2026intelligent}. 

\paragraph{Fire severity term.}
A per-instance fire severity score $H_i$ is constructed from segmentation confidence and size cues. Let $A^{px}_i$ be the fire mask area in pixels (or bounding-box area if masks are unavailable in a given implementation). Define a normalized size term
\begin{equation}
\tilde{A}_i=\frac{A^{px}_i}{A^{px}_{frame}},
\end{equation}
where $A^{px}_{frame}$ is the total frame area in pixels. A simple bounded severity form is
\begin{equation}
H_i=\sigma\!\left(\alpha_s\,s^F_i+\alpha_A\,\tilde{A}_i\right),
\end{equation}
where $\sigma(\cdot)$ is a logistic squashing function to keep $H_i\in(0,1)$, and $\alpha_s,\alpha_A\ge 0$ are weighting parameters. This choice is consistent with the paper’s emphasis that confidence and mask-level precision are critical for practical deployment and interpretability of fire segmentation outputs. :contentReference[oaicite:11]{index=11} :contentReference[oaicite:12]{index=12}

\paragraph{Object vulnerability term.}
Different surrounding entities imply different consequence levels. Let $V(c_j)\in[0,1]$ denote the vulnerability weight for object class $c_j$ (COCO category). For example, people may be assigned a high vulnerability, while static background objects may be assigned a lower vulnerability. To reflect object-detection reliability, incorporate the object confidence score:
\begin{equation}
C_j=\sigma\!\left(\beta_s\,s^O_j\right),
\end{equation}
with $\beta_s\ge 0$. The combined vulnerability-confidence factor becomes $V(c_j)\,C_j$.

\paragraph{Distance-to-exposure mapping.}
Exposure should increase as distance decreases. Using the metric distance $d^m_{ij}$, define an exposure function $E(d)$ that is high near the fire and decays with distance. A practical engineering form that avoids singularities is
\begin{equation}
E(d^m_{ij})=\exp\!\left(-\frac{d^m_{ij}}{\lambda}\right),
\end{equation}
where $\lambda>0$ sets the decay rate (interpretable as a characteristic distance beyond which exposure drops substantially). This is directly compatible with the paper’s proximity analysis concept, which distinguishes isolated low-risk fire from imminent threats to critical assets based on distance in meters. :contentReference[oaicite:13]{index=13} :contentReference[oaicite:14]{index=14}

\paragraph{Pairwise risk and frame-level aggregation.}
The pairwise risk contribution of fire instance $f_i$ to object $o_j$ is defined as
\begin{equation}
R_{ij}=H_i \cdot V(c_j)\cdot C_j \cdot E(d^m_{ij}).
\end{equation}
The object-level risk for a given object $o_j$ can be taken as the maximum threat over all fire instances:
\begin{equation}
R^{obj}_j=\max_{i\in\{1,\dots,N_F\}} R_{ij}.
\end{equation}
The frame-level risk score aggregates over objects of interest:
\begin{equation}
R^{frame}=\max_{j\in\{1,\dots,N_O\}} R^{obj}_j,
\end{equation}
or, if multiple simultaneous threats should accumulate, a bounded sum may be used:
\begin{equation}
R^{frame}=1-\prod_{j=1}^{N_O}\left(1-R^{obj}_j\right).
\end{equation}
These aggregation rules preserve interpretability (single worst-case threat) while allowing an alternative that captures multiple concurrent exposures.

\paragraph{Risk categorization and alert logic.}
To convert the continuous score into actionable outputs, define discrete risk tiers:
\begin{equation}
\text{Tier}(R^{frame})=
\begin{cases}
\text{Low}, & 0\le R^{frame}<\tau_1,\\
\text{Medium}, & \tau_1\le R^{frame}<\tau_2,\\
\text{High}, & \tau_2\le R^{frame}<\tau_3,\\
\text{Critical}, & \tau_3\le R^{frame}\le 1,
\end{cases}
\end{equation}
with thresholds $0<\tau_1<\tau_2<\tau_3<1$ selected based on operational tolerance. To align with the paper’s proximity-visualization workflow, an alert can be triggered when any object satisfies both a distance condition and a risk condition:
\begin{equation}
\exists (i,j)\ \text{s.t.}\ d^m_{ij}\le d_{crit}\ \land\ R_{ij}\ge \rho_{crit},
\end{equation}
where $d_{crit}$ is a safety radius (meters) and $\rho_{crit}$ is a minimum risk level for escalation. The annotated outputs described in the paper (bounding boxes, proximity lines, and distance labels) naturally support this logic by exposing the driving distances and object categories used in decision-making. :contentReference[oaicite:15]{index=15} :contentReference[oaicite:16]{index=16}

\paragraph{Temporal smoothing for video and animated sequences.}
Because the pipeline can process image sequences and generate annotated animations, risk should be stabilized across frames to reduce flicker and prevent single-frame outliers from dominating decisions. Let $R^{frame}_t$ be the instantaneous risk at time index $t$. A simple exponential smoothing filter is
\begin{equation}
\bar{R}_t=\gamma\,\bar{R}_{t-1}+(1-\gamma)\,R^{frame}_t,\qquad 0<\gamma<1,
\end{equation}
and alerts are issued on $\bar{R}_t$ rather than $R^{frame}_t$. This matches the paper’s emphasis on producing interpretable temporal outputs (annotated frames and saved sequences) that support real-time or post-event analysis. :contentReference[oaicite:17]{index=17} :contentReference[oaicite:18]{index=18}

\paragraph{Calibration and uncertainty handling.}
Given that metric distance accuracy depends on pixel-to-meter scaling and monocular projection effects, the risk model explicitly treats distance as approximate. A conservative safety choice is to downscale $\kappa$ (equivalently inflate distances less) to avoid underestimating risk, or to propagate distance uncertainty $\delta d$ into exposure by evaluating a worst-case bound:
\begin{equation}
E_{wc}(d^m_{ij})=\exp\!\left(-\frac{\max(d^m_{ij}-\delta d,\,0)}{\lambda}\right).
\end{equation}
This is consistent with the paper’s discussion that without camera calibration or reliable in-frame scale references, distance values should be interpreted as approximations affected by camera angle, lens distortion, resolution, and depth variations. :contentReference[oaicite:19]{index=19}

\paragraph{Implementation alignment with the proposed framework.}
Operationally, the model is computed per frame using the same intermediate outputs already produced by the dual-model pipeline: fire masks/bounding boxes and confidences from the custom YOLOv8 segmentation model; object boxes, labels, and confidences from the COCO-pretrained detector; centroid-based Euclidean distances; and pixel-to-meter conversion using a calibrated scaling factor. Therefore, the proposed risk score can be added without changing the detection backbone, and it directly complements the paper’s aim of transforming fire detection into proximity-aware hazard prioritization and decision support.

\section{Conclusion}\label{sec:conclusion}

This study presented an enhanced YOLOv8-based instance segmentation framework that advances conventional fire and smoke detection by coupling high-precision visual recognition with proximity-aware hazard reasoning and quantitative risk assessment. Trained for 200 epochs on a diverse annotated dataset, the proposed segmentation model demonstrated strong and consistent performance across standard evaluation metrics, achieving values above 90\% for mAP@0.5, precision, and F1 score, which indicates robust generalization across varied fire and smoke appearances. Building on this detection capability, the framework integrated a secondary COCO-pretrained object detection layer to identify surrounding entities of operational relevance, including people, vehicles, infrastructure, and equipment, thereby enabling contextual interpretation of detected fire events.

A key contribution of this work is the integration of spatial mapping with risk-aware decision logic. By estimating pairwise Euclidean distances between fire instances and nearby detected objects and converting pixel measurements into approximate metric distances through a pixels-per-meter scaling procedure, the system provides interpretable proximity indicators suitable for safety-oriented monitoring. These proximity measurements were then embedded into a structured risk assessment model that combines fire severity cues and detection confidence with object vulnerability and distance-based exposure decay to produce an actionable risk score and corresponding alert tiers. This extends the system from a detection-only tool into a prioritization mechanism that supports rapid assessment of which assets are most threatened and why, rather than reporting fire presence alone.

Through this dual-model, proximity-to-risk pipeline, the framework delivers actionable outputs that identify not only the location of fire or smoke, but also the relative threat to high-value or vulnerable targets in the scene. The resulting annotated frames and animated sequences provide an intuitive visualization of detections, distances, and risk context, supporting both real-time situational awareness and post-event analysis, and offering a practical pathway for integration into broader safety management and emergency response systems. Overall, the study demonstrates how modern deep learning can be extended from accurate segmentation toward interpretable, decision-supportive risk intelligence for industrial, engineering, and public safety environments.




\clearpage 

\begin{figure}
  \centering
    \caption{}\label{fig1}
\end{figure}




\section{}\label{}

\printcredits

\bibliographystyle{cas-model2-names}

\bibliography{cas-refs}



\end{document}